%% file: ms.tex
\pdfoutput=1

\documentclass[twocolumn, switch]{article} 
\input{sections/00-preamble}

\usepackage{scrextend}
\addtokomafont{footnote}{\fontsize{7pt}{9pt}\selectfont}
\deffootnote[0em]{1em}{0.1em}{\textsuperscript{\makebox[1em][l]{\thefootnotemark}}}

\usepackage{preprint}

\usepackage{amsmath, amsthm, amssymb, amsfonts}

\usepackage[numbers,square]{natbib}
\usepackage{textcomp}
\usepackage{xcolor}		
\usepackage[colorlinks = true,
            linkcolor = black,
            urlcolor  = blue,
            citecolor = cyan,
            anchorcolor = black]{hyperref}	
\usepackage{booktabs} 		
\usepackage{nicefrac}		
\usepackage{microtype}		
\usepackage{lineno}		
\usepackage{float}			

\usepackage{lipsum}		

\usepackage{newfloat}
\DeclareFloatingEnvironment[name={Supplementary Figure}]{suppfigure}
\usepackage{sidecap}
\sidecaptionvpos{figure}{c}

\usepackage{titlesec}
\titlespacing\section{0pt}{12pt plus 3pt minus 3pt}{1pt plus 1pt minus 1pt}
\titlespacing\subsection{0pt}{10pt plus 3pt minus 3pt}{1pt plus 1pt minus 1pt}
\titlespacing\subsubsection{0pt}{8pt plus 3pt minus 3pt}{1pt plus 1pt minus 1pt}

\title{Optuna:  A Next-generation Hyperparameter Optimization Framework}

\usepackage{xwatermark}
\newwatermark[firstpage,color=gray!60,angle=90,scale=0.32, xpos=3.9in,ypos=0]{\href{https://doi.org/}{\color{gray}{Preprint doi}}}

\usepackage{authblk}

\author[1\thanks{\tt{akiba@preferred.jp}}]{Takuya Akiba}
\author[1]{Shotaro Sano}
\author[1]{Toshihiko Yanase}
\author[1]{Takeru Ohta}
\author[1]{Masanori Koyama}
\affil[1]{Preferred Networks, Inc.}

\begin{document}

\twocolumn[ 
  \begin{@twocolumnfalse} 

\maketitle

\input{sections/01-abstract}
\vspace{0.35cm}

  \end{@twocolumnfalse} 
] 



\input{sections/02-introduction}

\input{sections/05-define_by_run}
\input{sections/06-optimization}
\input{sections/07-system}
\input{sections/08-evaluation}
\input{sections/09-case_study}
\input{sections/10-conclusions}

\bibliography{optuna}

\end{document}

%% file: sections/00-preamble.tex
\usepackage{url}            
\usepackage{booktabs}       
\usepackage{amsfonts}       
\usepackage{listings}
\usepackage{graphicx}
\usepackage{subfig}

\newcommand{\ignore}[1]{}

\usepackage[utf8]{inputenc}
\usepackage{setspace}

\DeclareFixedFont{\ttb}{T1}{txtt}{bx}{n}{7} 
\DeclareFixedFont{\ttm}{T1}{txtt}{m}{n}{7}  

\usepackage{color}
\definecolor{deepblue}{rgb}{0,0,0.5}
\definecolor{deepred}{rgb}{0.6,0,0}
\definecolor{deepgreen}{rgb}{0,0.5,0}

\usepackage{listings}
\usepackage[ruled,linesnumbered]{algorithm2e}
\usepackage[export]{adjustbox}

\newcommand\pythonstyle{\lstset{
language=Python,
breaklines=true,
numbers=left,
numberstyle=\tiny,
basicstyle=\setstretch{0.5}\ttm,  %
otherkeywords={self},             
keywordstyle=\ttb\color{deepblue},  
emphstyle=\ttb\color{deepred},    
stringstyle=\color{deepred},
commentstyle=\color{deepgreen},
frame=tb,                         
showstringspaces=false,            %
xleftmargin=2em,
framexleftmargin=2.5em
}}

\lstnewenvironment{python}[1][]
{
\pythonstyle
\lstset{#1}
}
{}

\newcommand\pythonexternal[2][]{{
\pythonstyle
\lstinputlisting[#1]{#2}}}

\def\backtick{\`{}}
\newcommand\bashstyle{\lstset{
language=bash,
breaklines=true,
numbers=left,
numberstyle=\tiny,
basicstyle=\setstretch{0.5}\ttm,  %
otherkeywords={self},             
keywordstyle=\ttb\color{deepblue},  
emphstyle=\ttb\color{deepred},    
stringstyle=\color{deepred},
commentstyle=\color{deepgreen},
frame=tb,                         
showstringspaces=false,            %
xleftmargin=2em,
framexleftmargin=2.5em,
upquote=true,
literate={`}{\backtick}1
}}

\lstnewenvironment{bash}[1][]
{
\bashstyle
\lstset{#1}
}
{}

\newcommand\bashexternal[2][]{{
\bashstyle
\lstinputlisting[#1]{#2}}}

\usepackage{pifont}
\newcommand{\cmark}{\ding{51}}%
\newcommand{\xmark}{\ding{55}}%

\newcommand{\myparagraph}[1]{\vspace{0.5em} \noindent {\bf #1.}}
\newcommand{\etal}[1]{{\it et al.}}


%% file: sections/01-abstract.tex
\begin{abstract}
The purpose of this study is to introduce new design-criteria for next-generation hyperparameter optimization software.
The criteria we propose include (1) \emph{define-by-run API} that allows users to construct the \emph{parameter search space} dynamically, (2) efficient implementation of both searching and pruning strategies, and (3) easy-to-setup, versatile architecture that can be deployed for various purposes, ranging from scalable distributed computing to light-weight experiment conducted via interactive interface.
In order to prove our point, we will introduce \emph{Optuna}, an optimization software which is a culmination of our effort in the development of a next generation optimization software.
As an optimization software designed with \emph{define-by-run} principle, \emph{Optuna} is particularly the first of its kind.
We will present the design-techniques that became necessary in the development of the software that meets the above criteria,
and demonstrate the power of our new design through experimental results and real world applications.
Our software is available under the MIT license (\url{https://github.com/pfnet/optuna/}). 
\end{abstract}

%% file: sections/02-introduction.tex
\section{Introduction}

\begin{sloppypar}  
Hyperparameter search is one of the most cumbersome tasks in machine learning projects.
The complexity of deep learning method is growing with its popularity, and the framework of efficient automatic hyperparameter tuning is in higher demand than ever.
Hyperparameter optimization softwares
such as \emph{Hyperopt}~\cite{1749-4699-8-1-014008},
\emph{Spearmint}~\cite{Snoek:2012:PBO:2999325.2999464},
\emph{SMAC}~\cite{Hutter:2011:SMO:2177360.2177404},
\emph{Autotune}~\cite{Koch2018autotune}, and
\emph{Vizier}~\cite{Golovin2017}
were all developed in order to meet this need.

The choice of the parameter-sampling algorithms varies across frameworks.
\emph{Spearmint}~\cite{Snoek:2012:PBO:2999325.2999464} and \emph{GPyOpt} use Gaussian Processes, and \emph{Hyperopt}~\cite{1749-4699-8-1-014008} employs \emph{tree-structured Parzen estimator} (TPE)~\cite{Bergstra:2011:AHO:2986459.2986743}.
Hutter~et~al. proposed \emph{SMAC}~\cite{Hutter:2011:SMO:2177360.2177404} that uses random forests.
Recent frameworks such as \emph{Google Vizier}~\cite{Golovin2017}, \emph{Katib} and \emph{Tune}~\cite{Liaw2018-Tune-hyperparameter-optimization} also support \emph{pruning} algorithms,
which monitor the intermediate result of each trial and kills the unpromising trials prematurely in order to  speed up the exploration.
There is an active research field for the pruning algorithm in hyperparameter optimization.
Domhan~et~al. proposed a method that uses parametric models to predict the learning curve~\cite{Domhan:2015:SUA:2832581.2832731}.
Klein~et~al. constructed Bayesian neural networks to predict the expected
learning curve~\cite{klein-iclr17}.
Li~et~al. employed a bandit-based algorithm and proposed Hyperband~\cite{Li2018-hyperband}.

A still another way to accelerate the optimization process is to use distributed computing, which enables parallel processing of multiple trials.
\emph{Katib} is built on \emph{Kubeflow}, which is a computing platform for machine learning services that is based on \emph{Kubernetes}.
\emph{Tune} also supports parallel optimization, and uses the \emph{Ray} distributed computing platform~\cite{moritz2017-ray}.
\end{sloppypar}

However, there are several serious problems that are being overlooked in many of these existing optimization frameworks.
Firstly, all previous hyperparameter optimization frameworks to date require the user to statically construct the parameter-search-space for each model,
and the search space can be extremely hard to describe in these frameworks for large-scale experiments that involve massive number of candidate models of different types with large parameter spaces and many conditional variables.
When the parameter space is not appropriately described by the user, application of advanced optimization method can be in vain.
Secondly, many existing frameworks do not feature efficient pruning strategy, when in fact
both \emph{parameter searching strategy} and \emph{performance estimation strategy} are important for high-performance optimization under limited resource availability \cite{automl}\cite{Golovin2017}\cite{Liaw2018-Tune-hyperparameter-optimization}.
Finally, in order to accommodate with a variety of models in a variety of situations, the architecture shall be able to handle both small and large scale experiments with minimum setup requirements.
If possible, architecture shall be installable with a single command as well, and it shall be designed as an open source software so that it can continuously incorporate newest species of optimization methods by interacting with open source community.

In order to address these concerns, we propose to introduce the following new design criteria for next-generation optimization framework:
\begin{itemize}
    \item  \emph{Define-by-run} programming that allows the user to dynamically construct the search space,
    \item Efficient sampling algorithm and pruning algorithm that allows some user-customization,
    \item  Easy-to-setup, versatile architecture that can be deployed for tasks of various types, ranging from light-weight experiments conducted via interactive interfaces to heavy-weight distributed computations.
\end{itemize}
In this study, we will demonstrate the significance of these criteria through \emph{Optuna}, an open-source optimization software which is a culmination of our effort in making our definition of next-generation optimization framework come to reality.

We will also present new design techniques and new optimization algorithms that we had to develop in order to meet our proposed criteria.
Thanks to these new design techniques, our implementation outperforms many major black-box optimization frameworks while being easy to use and easy to setup in various environments.
In what follows, we will elaborate each of our proposed criteria together with our technical solutions, and present experimental results in both real world applications and benchmark datasets.

\emph{Optuna} is released under the MIT license (\url{https://github.com/pfnet/optuna/}), and is in production use at Preferred Networks for more than one year.
\input{tables/frameworks_api_styles}

\vspace{0.5cm}

%% file: tables/frameworks_api_styles.tex
\newcommand{\y}[1]{{\small{(#1)}}}

\begin{table*}[t]
    \centering
    \caption{Software frameworks for deep learning and hyperparameter optimization,
    sorted by their API styles: \emph{define-and-run} and \emph{define-by-run}.
    }
    \fontsize{8pt}{10pt}\selectfont
      \tabcolsep=3.5mm
    \scalebox{0.82}{
    \begin{tabular}{c||c|c}
        \toprule
         & \textbf{Deep Learning Frameworks} & \textbf{Hyperparameter Optimization Frameworks} \\
        \midrule \midrule
            \begin{tabular}{@{}c@{}}\textbf{Define-and-Run Style} \\ {\small (symbolic, static)} \end{tabular}
        &
            \begin{tabular}{@{}c@{}}Torch \y{2002}, Theano \y{2007}, Caffe \y{2013}, \\  TensorFlow \y{2015}, MXNet \y{2015}, Keras \y{2015}\end{tabular}
        &
            \begin{tabular}{@{}c@{}}SMAC \y{2011}, Spearmint \y{2012}, Hyperopt \y{2015}, GPyOpt \y{2016},  \\  Vizier \y{2017}, Katib \y{2018}, Tune \y{2018}, Autotune \y{2018}\end{tabular} \\
        \midrule
            \begin{tabular}{@{}c@{}}\textbf{Define-by-Run Style} \\ {\small (imperative, dynamic)} \end{tabular}
        &
            \begin{tabular}{@{}c@{}}Chainer \y{2015}, DyNet \y{2016}, PyTorch \y{2016}, \\ TensorFlow Eager \y{2017}, Gluon \y{2017}\end{tabular}
        &
            \textit{\textbf{Optuna \y{2019; this work}}} \\
        \bottomrule
    \end{tabular}
    }
    \label{tbl:frameworks_api_styles}
\end{table*}

%% file: sections/05-define_by_run.tex
\begin{figure}[t]
            \pythonexternal{codes/optuna_sklearn_nn.txt}
            \caption{
            An example code of \emph{Optuna}'s \emph{define-by-run} style API.
            This code builds a space of hyperparameters for a classifier of the MNIST dataset and optimizes the number of layers and the number of hidden units at each layer.
            }
            \label{fig:optuna_sklearn}
\end{figure}

\begin{figure}[t]
    \pythonexternal{codes/hyperopt_sklearn_nn.txt}
    \caption{
    An example code of \textit{Hyperopt}~\cite{1749-4699-8-1-014008} that has the exactly same functionality as the code in \ref{fig:optuna_sklearn}.  \textit{Hyperopt} is an example of \emph{define-and-run} style API.}
    \label{fig:Hyperopt}
\end{figure}

\section{Define-by-run API}
In this section we describe the significance of the \emph{define-by-run} principle.
As we will elaborate later, we are borrowing the term \emph{define-by-run} from a trending philosophy in deep learning frameworks that allows the user to dynamically program deep networks.
Following the original definition,  we use the term \emph{define-by-run} in the context of optimization framework to refer to a design that allows the user to dynamically construct the search space.
In \emph{define-by-run API}, the user does not have to bear the full burden of explicitly defining everything in advance about the optimization strategy.

The power of \emph{define-by-run} API is more easily understood with actual code.
\emph{Optuna} formulates the hyperparameter optimization as a process of minimizing/maximizing an \emph{objective function} that takes a set of hyperparameters as an input and returns its (validation) score.
Figure~\ref{fig:optuna_sklearn} is an example of an objective function written in \emph{Optuna}.
This function dynamically constructs the search space of neural network architecture (the number of layers and the number of hidden units) without relying on externally defined static variables.
\emph{Optuna} refers to each process of  optimization as a \emph{study}, and to each evaluation of objective function as a \emph{trial}.
In the code of Figure~\ref{fig:optuna_sklearn}, \emph{Optuna} defines an objective function  (Lines~4--18), and invokes the `\texttt{optimize API}' that takes the objective function as an input (Line~21).
Instead of hyperparameter values, an objective function in \emph{Optuna} receives a \emph{living trial object}, which is associated with a single \emph{trial}.

\emph{Optuna} gradually builds the objective function through the interaction with the \emph{trial} object.
The search spaces are constructed dynamically by the methods of the \emph{trial} object during the runtime of the objective function.
The user is asked to invoke
`{\texttt suggest API}'  inside the objective function in order to dynamically generate the hyperparameters for each \emph{trial} (Lines~5 and 9).
Upon the invocation of `\texttt{suggest API}', a hyperparameter is statistically sampled based on the history of previously evaluated \emph{trials}.
At Line~5, `\texttt{suggest\_int}' method suggests a value for `\texttt{n\_layers}', the integer hyperparameter that determines the number of layers in the Multilayer Perceptron.
Using loops and conditional statements written in usual \emph{Python} syntax, the user can easily represent a wide variety of parameter spaces.
With this functionality, the user of \emph{Optuna} can even express heterogeneous parameter space with an intuitive and simple code (Figure~\ref{fig:hetero}).

Meanwhile, Figure~\ref{fig:Hyperopt} is an example code of \emph{Hyperopt} that has the exactly same functionality as the \textit{Optuna} code in Figure~\ref{fig:optuna_sklearn}.
Note that the same function written in \emph{Hyperopt} (Figure~\ref{fig:Hyperopt}) is significantly longer, more convoluted, and harder to interpret.
It is not even obvious at first glance that the code in Figure~\ref{fig:Hyperopt} is in fact equivalent to the code in Figure~\ref{fig:optuna_sklearn}!
In order to write the same \emph{for-loop} in Figure~\ref{fig:optuna_sklearn} using \emph{Hyperopt}, the user must prepare the list of all the parameters in the parameter-space prior to the exploration (see line 4-18 in Figure~\ref{fig:Hyperopt}).
This requirement will lead the user to even darker nightmares when the optimization problem is more complicated.

\input{tables/related_work} 

\subsection{Modular Programming}
\label{sec:define-by-run:drawback}

\begin{figure}
\pythonexternal{codes/modularity.txt}
\caption{An example code of \emph{Optuna} for the construction of a heterogeneous parameter-space. This code simultaneously explores the parameter spaces of both random forest and MLP.}
\label{fig:hetero}
\end{figure}

\begin{figure}
\pythonexternal{codes/optuna_mnist_clipped.txt}
\caption{
Another example of \emph{Optuna}'s objective function.
This code simultaneously optimizes
neural network architecture (the \texttt{create\_model} method)
and the hyperparameters for stochastic gradient descent (the \texttt{create\_optimizer} method).
}
\label{fig:optuna_mnist}
\end{figure}

\begin{sloppypar}
A keen reader might have noticed in Figure \ref{fig:hetero} that the optimization code written in \emph{Optuna} is highly modular, thanks to its \emph{define-by-run} design.
Compatibility with modular programming is another important strength of the \emph{define-by-run} design.
Figure~\ref{fig:optuna_mnist} is another example code written in \emph{Optuna}
for a more complex scenario.
This code is capable of simultaneously optimizing both the topology of a multilayer perceptron (method `\texttt{create\_model}')  and the hyperparameters of stochastic gradient descent (method `\texttt{create\_optimizer}').
The method `\texttt{create\_model}' generates `\texttt{n\_layers}' in Line~5 and uses a \emph{for loop} to construct a neural network of depth equal to `\texttt{n\_layers}'.
The method also generates `\texttt{n\_units\_$i$}' at each $i$-th loop, a hyperparameter that determines the number of the units in the $i$-th layer.
The method `\texttt{create\_optimizer}', on the other hand, makes suggestions for both learning rate and weight-decay parameter.
Again,  a complex space of hyperparameters is simply expressed in \emph{Optuna.}
Most notably, in this example, the methods `\texttt{create\_model}' and `\texttt{create\_optimizer}' are independent of one another, so that we can make changes to each one of them separately.
Thus, the user can easily augment this code with other conditional variables and methods for other set of parameters, and make a choice from more diverse pool of models.
\end{sloppypar}

\subsection{Deployment}

Indeed, the benefit of our \emph{define-by-run} API means nothing if we cannot easily deploy the model with the best set of hyperparameters found by the algorithm.
The above example (Figure~\ref{fig:optuna_mnist}) might make it seem as if the user has to write a different version of the objective function that does not invoke `\texttt{trial.suggest}' in order to deploy the \textit{best} configuration.
Luckily, this is not a concern.
For deployment purpose, \emph{Optuna} features a separate class called `\texttt{FixedTrial}' that can be passed to objective functions.
The `\texttt{FixedTrial}' object has practically the same set of functionalities as the \emph{trial} class, except that it will only suggest the user defined set of the hyperparameters  when passed to the objective functions.
Once a parameter-set of interest is found (e.g., the best ones), the user simply has to construct a `\texttt{FixedTrial}' object with the parameter set.

\subsection{Historical Remarks}
Historically, the term \emph{define-by-run} was coined by the developers of deep learning frameworks.
In the beginning, most deep learning frameworks like \emph{Theano} and \emph{Torch} used to be \textit{declarative}, and constructed the networks in their \textit{domain specific languages} (DSL).
These frameworks are called \emph{define-and-run} frameworks because they do not allow the user to alter the manipulation of intermediate variables once the network is defined.
In \emph{define-and-run} frameworks, computation is conducted in two phases: (1) construction phase and (2) evaluation phase.
In a way, contemporary optimization methods like \emph{Hyperopt} are built on the philosophy similar to \emph{define-and-run}, because there are two phases in their optimization: (1) construction of the search space and (3) exploration in the search space.

Because of their difficulty of programming, the \emph{define-and-run} style deep learning frameworks are quickly being replaced by \emph{define-by-run} style deep learning frameworks like \emph{Chainer}~\cite{chainer_learningsys2015},
\emph{DyNet}~\cite{dynet}, \emph{PyTorch}~\cite{pytorch}, eager-mode \emph{TensorFlow}~\cite{tensorflow}, and \emph{Gluon}.
In the \emph{define-by-run} style DL framework, there are no two separate phases for the construction of the network and the computation on the network.
Instead, the user is allowed to directly program how each variables are to be manipulated in the network.
What we propose in this article is an analogue of the \emph{define-by-run} DL framework for hyperparameter optimization, in which the framework asks the user to directly program the \emph{parameter search-space}~(See Table \ref{tbl:frameworks_api_styles}) .
Armed with the architecture built on the \emph{define-by-run} principle, our \emph{Optuna} can express highly sophisticated search space at ease.

%% file: tables/related_work.tex
\newcommand{\K}{\textcolor{blue}{\cmark}}
\newcommand{\X}{\textcolor{red}{\xmark}}
\newcommand{\N}{\textbf{--}}

\begin{table*}
\caption{
Comparison of previous hyperparameter optimization frameworks and \emph{Optuna}. There is a checkmark for \emph{lightweight} if the setup for the framework is easy and it can be easily used for lightweight purposes.
}
\centering
\fontsize{8pt}{10pt}\selectfont
\tabcolsep=3.5mm
\begin{tabular}{c|ccccccc}
\toprule
\textbf{Framework} & \textbf{API Style} & \textbf{Pruning} & \textbf{Lightweight} & \textbf{Distributed} & \textbf{Dashboard} & \textbf{OSS} \\
\midrule
SMAC~\cite{Hutter:2011:SMO:2177360.2177404}     & define-and-run & \X&\K&\X&\X&\K \\
GPyOpt & define-and-run & \X&\K&\X&\X&\K \\
Spearmint~\cite{Snoek:2012:PBO:2999325.2999464} & define-and-run & \X&\K&\K&\X&\K \\
Hyperopt~\cite{1749-4699-8-1-014008}            & define-and-run & \X&\K&\K&\X&\K \\
Autotune~\cite{Koch2018autotune}                       & define-and-run & \K&\X&\K&\K&\X \\
Vizier~\cite{Golovin2017}                       & define-and-run & \K&\X&\K&\K&\X \\
Katib                         & define-and-run & \K&\X&\K&\K&\K \\
Tune~\cite{Liaw2018-Tune-hyperparameter-optimization}        & define-and-run &\K&\X&\K&\K&\K \\
\midrule
Optuna {\tiny (this work)}                      & define-by-run  &\K&\K&\K&\K&\K \\
\bottomrule
\end{tabular}
\label{tbl:oss-comparison}
\end{table*}

%% file: sections/06-optimization.tex
\section{Efficient Sampling and Pruning Mechanism}
In general, the cost-effectiveness of hyperparameter optimization framework is determined by the efficiency of (1) \textit{searching strategy} that determines the set of parameters that shall be investigated, and (2) \textit{performance estimation strategy }that estimates the value of currently investigated parameters from learning curves and determines the set of parameters that shall be discarded.
As we will experimentally show later, the efficiency of both
searching strategy and performance estimation strategy is necessary for cost-effective optimization method.

The strategy for the termination of unpromising \emph{trials} is often referred to as \emph{pruning} in many literatures, and
it is also well known as {\it automated early stopping}~\cite{Golovin2017}~\cite{Liaw2018-Tune-hyperparameter-optimization}. We, however, refer to this functionality as {\it pruning} in order to distinguish it from the \textit{early stopping regularization} in machine learning that exists as a countermeasure against overfitting.
As shown in table \ref{tbl:oss-comparison}, many existing frameworks do not provide efficient pruning strategies.
In this section we will provide our design for both \emph{sampling} and \emph{pruning}.

\subsection{Sampling Methods on Dynamically Constructed Parameter Space}

There are generally two types of sampling method: \emph{relational sampling} that exploits the correlations among the parameters and \emph{independent sampling} that samples each parameter independently.
The \emph{independent sampling} is not necessarily a naive option, because some sampling algorithms like TPE~\cite{Bergstra:2011:AHO:2986459.2986743} are known to perform well even without using the parameter correlations, and the cost effectiveness for both relational and independent sampling depends on environment and task.
Our \emph{Optuna} features both, and it can handle various independent sampling methods including TPE as well as relational sampling methods like CMA-ES.
However, some words of caution are in order for the implementation of \emph{relational sampling} in \emph{define-by-run} framework.

\subsubsection*{Relational sampling in define-by-run frameworks}
One valid claim about the advantage of the old  \emph{define-and-run} optimization design is that the program is given the knowledge of the concurrence relations among the hyperparamters from the beginning of the optimization process.
Implementing of optimization methods that takes the concurrence relations among the parameters into account is a nontrivial challenge when the search spaces are dynamically constructed.
To overcome this challenge, \emph{Optuna} features an ability to identify \emph{trial} results that are informative about the concurrence relations.
This way, the framework can identify the underlying concurrence relations after some number of independent samplings, and use the inferred concurrence relation to conduct user-selected relational sampling algorithms like CMA-ES~\cite{Hansen-cmaes:doi:10.1162/106365601750190398} and GP-BO~\cite{shahriari2016taking}.
Being an open source software, \emph{Optuna} also allows the user to use his/her own customized sampling procedure.

\subsection{Efficient Pruning Algorithm}

\begin{figure}[t]
\begin{python}
import ...

def objective(trial):
    ...

    lr = trial.suggest_loguniform('lr', 1e-5, 1e-1)
    clf = sklearn.linear_model.SGDClassifier(learning_rate=lr)
    for step in range(100):
        clf.partial_fit(x_train, y_train, classes)

        # Report intermediate objective value.
        intermediate_value = clf.score(x_val, y_val)
        trial.report(intermediate_value, step=step)

        # Handle pruning based on the intermediate value.
        if trial.should_prune(step):
            raise TrialPruned()

    return 1.0 - clf.score(x_val, y_val)

study = optuna.create_study()
study.optimize(objective)
\end{python}
\caption{An example of implementation of a pruning algorithm with \emph{Optuna}. An intermediate value is reported at each step of iterative training. The \texttt{Pruner} class stops unpromising \emph{trials} based on the history of reported values.}
\label{pruning fig}
\end{figure}

Pruning algorithm is essential in ensuring the "cost" part of the cost-effectiveness.
Pruning mechanism in general works in two phases. It \emph{(1)} periodically monitors the intermediate objective values, and \emph{(2)} terminates the \emph{trial} that does not meet the predefined condition.
In \emph{Optuna}, `\texttt{report API}' is responsible for the monitoring functionality, and `\texttt{should\_prune API}' is responsible for the premature termination of the unpromising \emph{trials} (see Figure~\ref{pruning fig}).
The background algorithm of `\texttt{should\_prune}' method is implemented by the family of \emph{pruner} classes.
\emph{Optuna} features a variant of Asynchronous Successive Halving algorithm \cite{Li2018massively} , a recently developed \textit{state of the art} method that scales linearly with the number of workers in distributed environment.

Asynchronous Successive Halving(ASHA) is an extension of Successive Halving~\cite{Jamieson2016non} in which each worker is allowed to asynchronously execute aggressive early stopping based on provisional ranking of \emph{trials}.
The most prominent advantage of asynchronous pruning is that it is particularly well suited for applications in distributional environment; because each worker does not have to wait for the results from other workers at each round of the pruning, the parallel computation can process multiple \emph{trials} simultaneously without delay.

Algorithm \ref{alg:asha_pruning} is the actual pruning algorithm implemented in \emph{Optuna}.
Inputs to the algorithm include the \emph{trial} that is subject to pruning, number of steps, reducing factor, minimum resource to be used before the pruning, and minimum early stopping rate.
Algorithm begins by computing the current \emph{rung} for the \emph{trial}, which is the number of times the \emph{trial} has survived the pruning.
The \emph{trial} is allowed to enter the next round of the competition if its provisional ranking is within top $1/ \eta$.
If the number of \emph{trials} with the same rung is less than $\eta$, the best \emph{trial} among the \emph{trials} with the same \emph{rung} becomes promoted.
In order to avoid having to record massive number of checkpointed configurations(snapshots), our implementation does not allow repechage.
As experimentally verify in the next section,  our modified implementation of Successive Halving scales linearly with the number of workers without any problem.
We will present the details of our optimization performance in  Section~\ref{Performance Evaluation of Pruning}.

\begin{algorithm}[t]
\KwIn{target trial $\mathtt{trial}$, current step $\mathtt{step}$, minimum resource $r$, reduction factor $\eta$, minimum early-stopping rate $s$.}
\KwOut{{\bf true} if the trial should be pruned, {\bf false} otherwise.}

$\mathtt{rung} \gets \max(0, \log{\eta}(\lfloor \mathtt{step}/r \rfloor) - s)$

\If{$\mathtt{step} \neq r\eta^{s+\mathtt{rung}}$} {
  \Return{\bf false}
}

$\mathtt{value} \gets $get\_trial\_intermediate\_value$(\mathtt{trial}, \mathtt{step})$

$\mathtt{values} \gets $get\_all\_trials\_intermediate\_values$(\mathtt{step})$

$\mathtt{top\_k\_values} \gets $top\_k$(\mathtt{values}, \lfloor |\mathtt{values}| / \eta \rfloor)$

\If{$\mathtt{top\_k\_values} = \emptyset$} {
  $\mathtt{top\_k\_values} \gets $top\_k$(\mathtt{values}, 1)$
}

\Return{$\mathtt{value} \notin \mathtt{top\_k\_values}$}

\caption{Pruning algorithm based on Successive Halving} \label{alg:asha_pruning}

\end{algorithm}

%% file: sections/07-system.tex
\section{Scalable and versatile System that is Easy to setup}

\begin{figure}
	  \begin{center}
          \includegraphics[width=7.5cm]{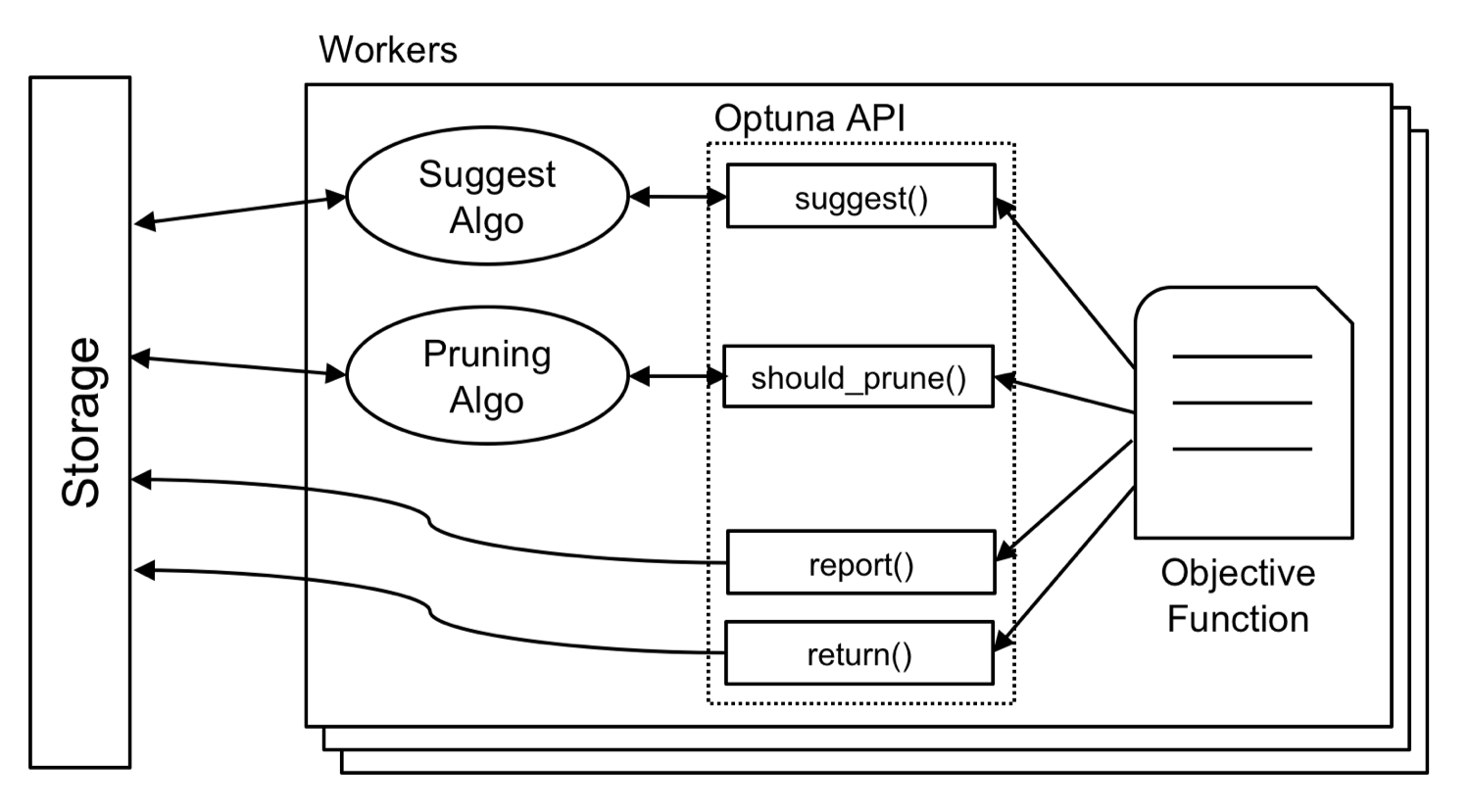}
          \captionof{figure}{Overview of \emph{Optuna}'s system design. Each worker executes one instance of an objective function in each \emph{study}.
          The Objective function runs its \emph{trial} using \emph{Optuna} APIs.
          When the API is invoked, the objective function accesses the shared storage and obtains the information of the past \emph{studies} from the storage when necessary.
          Each worker runs the objective function independently and shares the progress of the current \emph{study} via the storage.
          }
          \label{fig:system}
	  \end{center}
\end{figure}

Our last criterion for the next generation optimization software is a scalable system that can handle a wide variety of tasks, ranging from a heavy experiment that requires a massive number of workers to a trial-level, light-weight computation conducted through interactive interfaces like \emph{Jupyter Notebook}.
The figure \ref{fig:system} illustrates how the database(\emph{storage}) is incorporated into the system of \emph{Optuna}; the \textit{trial} objects shares the evaluations history of objective functions via storage.
\emph{Optuna} features a mechanism that allows the user to change the storage backend to meet his/her need.

\begin{figure}[t]
\centering
	\includegraphics[width=8cm]{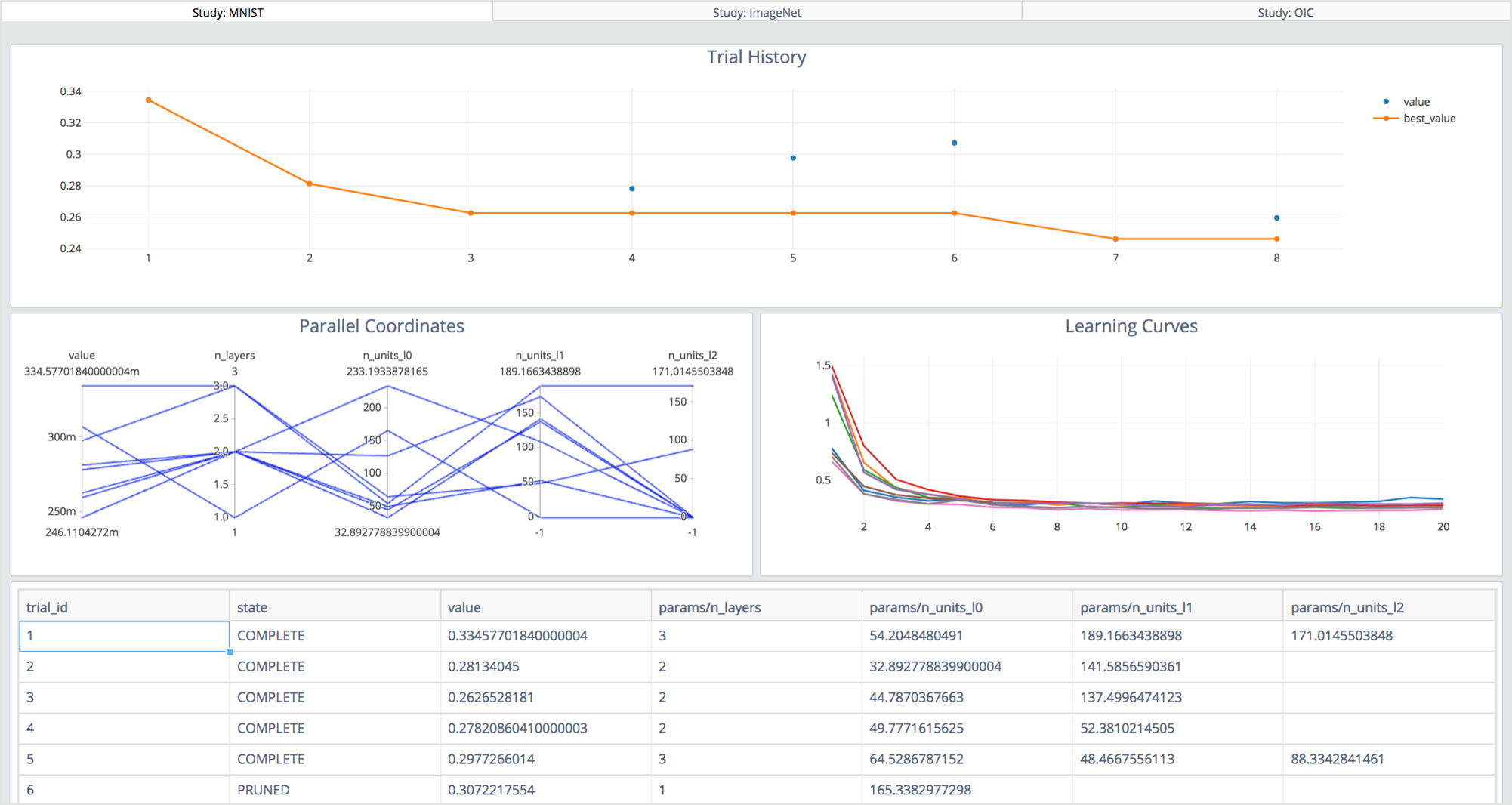}
	\captionof{figure}{\emph{Optuna} dashboard. This example shows the online transition of objective values, the parallel coordinates plot of sampled parameters, the learning curves, and the tabular descriptions of investigated \textit{trials}.}
	\label{fig:dashboard}
\end{figure}

\begin{sloppypar}
For example, when the user wants to run experiment with \emph{Jupyter Notebook} in a local machine, the
user may want to avoid spending effort in accessing a multi-tenant system deployed by some organization or in deploying a database on his/her own.
When there is no specification given, \emph{Optuna} automatically uses its built-in in-memory data-structure as the storage back-end.
From general user's perspective, that the framework can be easily used for lightweight purposes is one of the most essential strengths of \emph{Optuna}, and it is a particularly important part of our criteria for next-generation optimization frameworks.
This \emph{lightweight purpose compatibility} is also featured by select few frameworks like \emph{Hyperopt} and \emph{GPyOt} as well.
The user of \emph{Optuna} can also conduct more involved analysis by exporting the results in the \emph{pandas}~\cite{pandas} dataframe, which is highly compatible with interactive analysis frameworks like \emph{Jupyter Notebooks}~\cite{Kluyver:2016aa}.
\emph{Optuna} also provides web-dashboard for visualization and analysis of  \emph{studies} in real time (see Figure \ref{fig:dashboard}).
\end{sloppypar}

Meanwhile, when the user wants to conduct distributed computation, the user of \emph{Optuna} can deploy relational database as the backend.
The user of \emph{Optuna} can also use \emph{SQLite} database as well.
The figure \ref{fig:parallel_example_shell} is an example code that deploys \emph{SQLite} database.
This code conducts distributed computation by simply executing run.py multiple times with the same \emph{study} identifier and the same storage URL.

\begin{figure}
\centering
\subfloat[][Python code: run.py]{
    \parbox[]{\columnwidth}{
            \pythonexternal{codes/parallel_python.txt}
            \label{fig:parallel_example_python}
    }
} \\

\subfloat[][Shell]{
    \parbox[]{\columnwidth}{
            \bashexternal{codes/parallel_shell.txt}
            \label{fig:parallel_example_shell}
    }
}
\caption{Distributed optimization in \emph{Optuna}. Figure (a) is the optimization script executed by one worker. Figure (b) is an example \emph{shell} for the optimization with multiple workers in a distributed environment.}
\label{fig:parallel_example}
\end{figure}

\emph{Optuna}'s new design thus significantly reduces the effort required for storage deployment.
This new design can be easily incorporated into a container-orchestration system like \emph{Kubernetes} as well.
As we verify in the experiment section, the distributed computations conducted with our flexible system-design scales linearly with the number of workers.
\emph{Optuna} is also an open source software that can be installed to user's system with one command.

%% file: sections/08-evaluation.tex
\section{Experimental Evaluation}

We demonstrate the efficiency of the new design-framework through three sets of  experiments.

\subsection{Performance Evaluation Using a Collection of Tests}

As described in the previous section, \emph{Optuna} not only allows the user to use his/her own customized sampling procedure that suits the purpose, but also comes with multiple built-in optimization algorithms including the mixture of independent and relational sampling, which is not featured in currently existing frameworks.
For example, \emph{Optuna} can use the mixture of TPE and CMA-ES.
We compared the optimization performance of the TPE+CMA-ES against those of other sampling algorithms on a collection of tests for black-box optimization~\cite{evalset2016, Dewancker2016}, which contains 56 test cases.
We implemented four adversaries to compare against TPE+CMA-ES: random search as a baseline method, \emph{Hyperopt}~\cite{1749-4699-8-1-014008} as a TPE-based method, \emph{SMAC3}~\cite{Hutter:2011:SMO:2177360.2177404} as a random-forest based method, and \emph{GPyOpt} as a Gaussian Process based method.
For TPE+CMA-ES,  we used TPE for the first 40 steps and used CMA-ES for the rest.
For the evaluation metric, we used the best-attained objective  value found in  80 \textit{trials}.
Following the work of Dewancker et al.~\cite{Dewancker2016}, we repeated each \emph{study} 30 times for each algorithm and applied Paired Mann-Whitney U test with $ \alpha = 0.0005 $ to the results in order to statistically compare TPE+CMA-ES's performance against the rival algorithms.

\begin{figure}
	  \begin{center}
			\includegraphics[width=6.5cm]{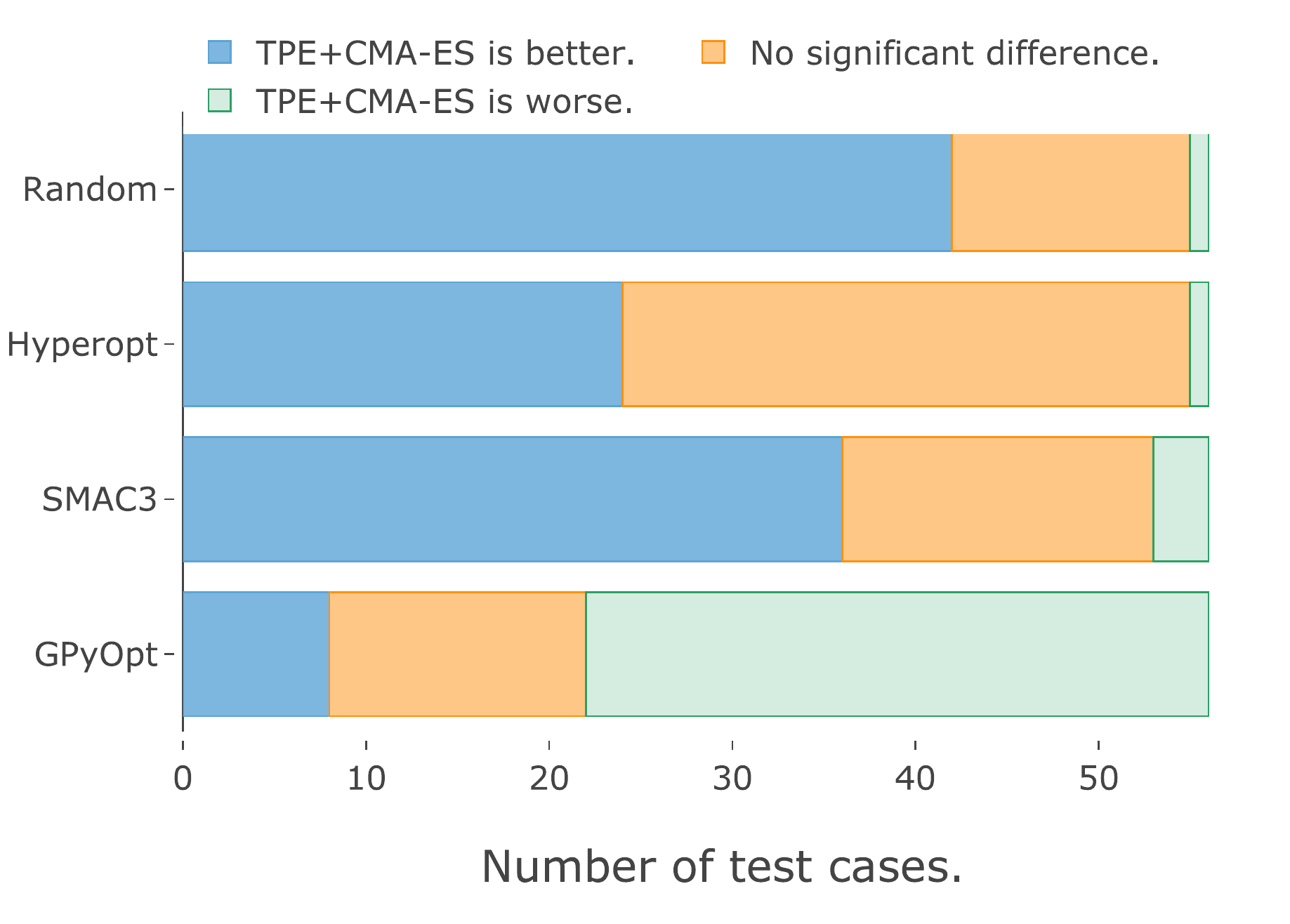}
	        \captionof{figure}{Result of comparing TPE+CMA-ES against other existing methods in terms of best attained objective value.  Each algorithm was applied to each \emph{study} 30 times, and Paired Mann-Whitney U test with $ \alpha = 0.0005 $ was used to determine whether TPE+CMA-ES outperforms each rival.}
	        \label{fig:results-test-collection}
	  \end{center}
\end{figure}

\begin{figure}
	  \begin{center}
			\includegraphics[width=6.5cm]{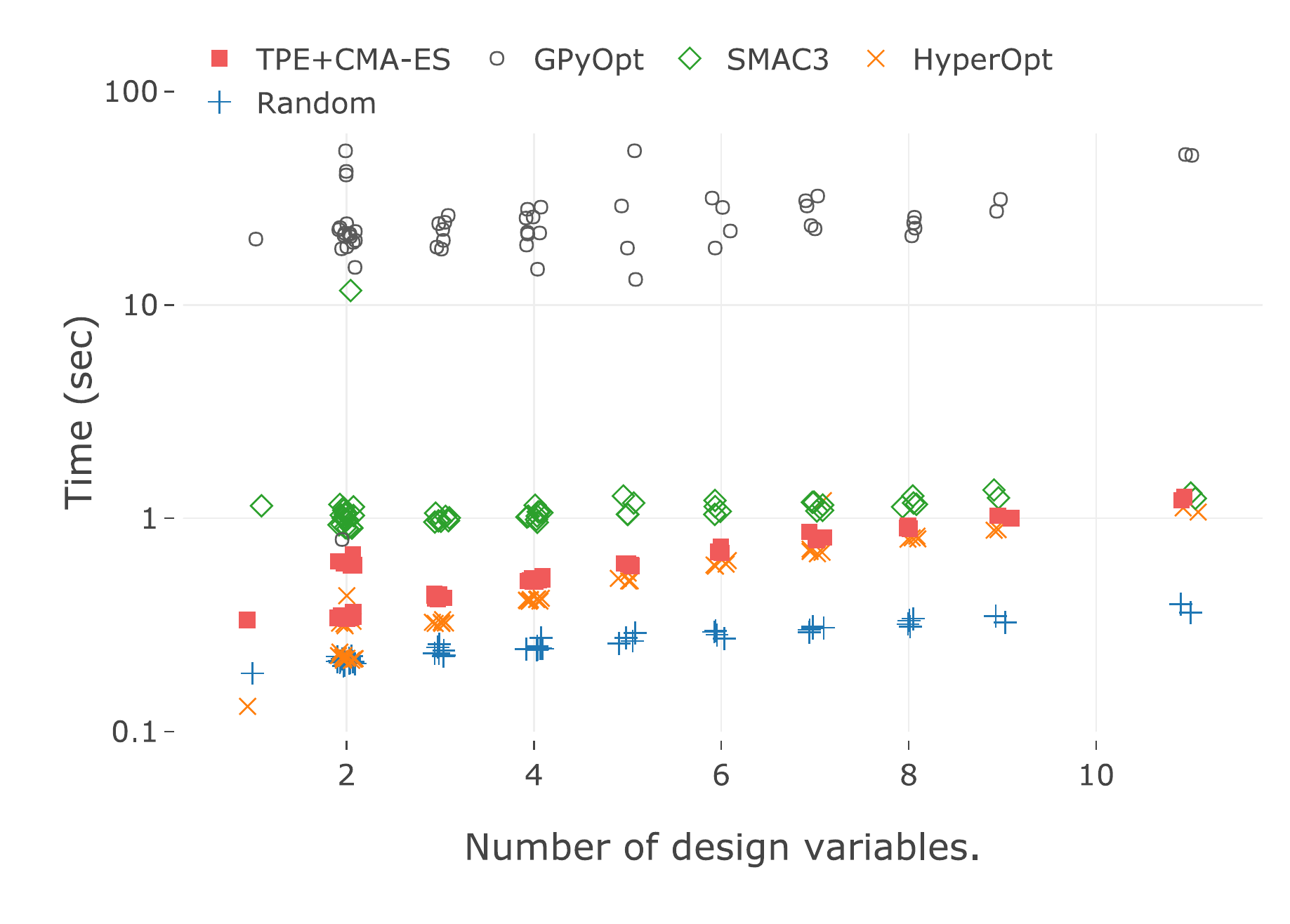}
	        \captionof{figure}{Computational time spent by different frameworks for each test case.	        }
	        \label{fig:test_collection_dim_vs_time}
	  \end{center}
\end{figure}

The results are shown in Figure~\ref{fig:results-test-collection}. TPE+CMA-ES finds statistically worse solution than random search in only 1/56 test cases, performs worse than \emph{Hyperopt} in 1/56 cases, and performs worse than \emph{SMAC3} in 3/56 cases.
Meanwhile, \emph{GPyOpt} performed better than TPE+CMA-ES in 34/56 cases in terms of the best-attained loss value.
At the same time, TPE+CMA-ES takes an order-of-magnitude less times per trial than \emph{GPyOpt}.
\begin{sloppypar}
Figure~\ref{fig:test_collection_dim_vs_time} shows the average time spent for each test case.
TPE+CMA-ES, \emph{Hyperopt}, \emph{SMAC3}, and random search finished one \textit{study} within few seconds even for the test case with more than ten design variables. On the other hand, \emph{GPyOpt} required twenty times longer duration to complete a \textit{study}.
We see that the mixture of TPE and CMA-ES is a cost-effective choice among current lines of advanced optimization algorithms.
If the time of evaluation is a bottleneck, the user may use Gaussian Process based method as a sampling algorithm.
We plan in near future to also develop an interface on which the user of \emph{Optuna} can easily deploy external optimization software as well.

\end{sloppypar}


\subsection{Performance Evaluation of Pruning} \label{Performance Evaluation of Pruning}

We evaluated the performance gain from the pruning procedure in the \emph{Optuna}-implemented optimization of Alex Krizhevsky's neural network (AlexNet)~\cite{Krizhevsky:2012:ICD:2999134.2999257} on the Street View House Numbers (SVHN) dataset~\cite{Netzer-2011-SVHN}.
We tested our pruning system together with random search and TPE.
Following the experiment in \cite{Li2018-hyperband}, we used a subnetwork of AlexNet (hereinafter called simplified AlexNet), which consists of three convolutional layers and a fully-connected layer and involves 8 hyperparameters.

For each experiment, we executed a \emph{study} with one NVIDIA Tesla P100 card, and terminated each \emph{study} $4$ hours into the experiment.
We repeated each \emph{study} 40 times.
With pruning, both TPE and random search was able to conduct a greater number of \emph{trials} within the same time limit.
On average, TPE and random search \emph{without} pruning completed 35.8 and 36.0 \emph{trials} per \emph{study}, respectively.
On the other hand, TPE \emph{with} pruning explored 1278.6 \emph{trials} on average per \emph{study}, of which 1271.5 were pruned during the process.
Random search \emph{with} pruning explored 1119.3 \emph{trials} with 1111.3 pruned \emph{trials}.

Figure \ref{fig:alexnet_svhn} shows the transition of the average test errors.
The result clearly suggests that pruning can significantly accelerate the optimization for both TPE and random search.
Our implementation of ASHA significantly outperforms Median pruning, a pruning method featured in \emph{Vizier}.
This result also suggests that sampling algorithm alone is not sufficient for cost-effective optimization.
The bottleneck of sampling algorithm is the computational cost required for each \emph{trial}, and pruning algorithm is necessary for fast optimization.

\subsection{Performance Evaluation of Distributed Optimization} \label{Performance Evaluation of Distributed Optimization}
\begin{figure*}[htb]
\centering
\subfloat[][]{
\includegraphics[width=5.7cm]{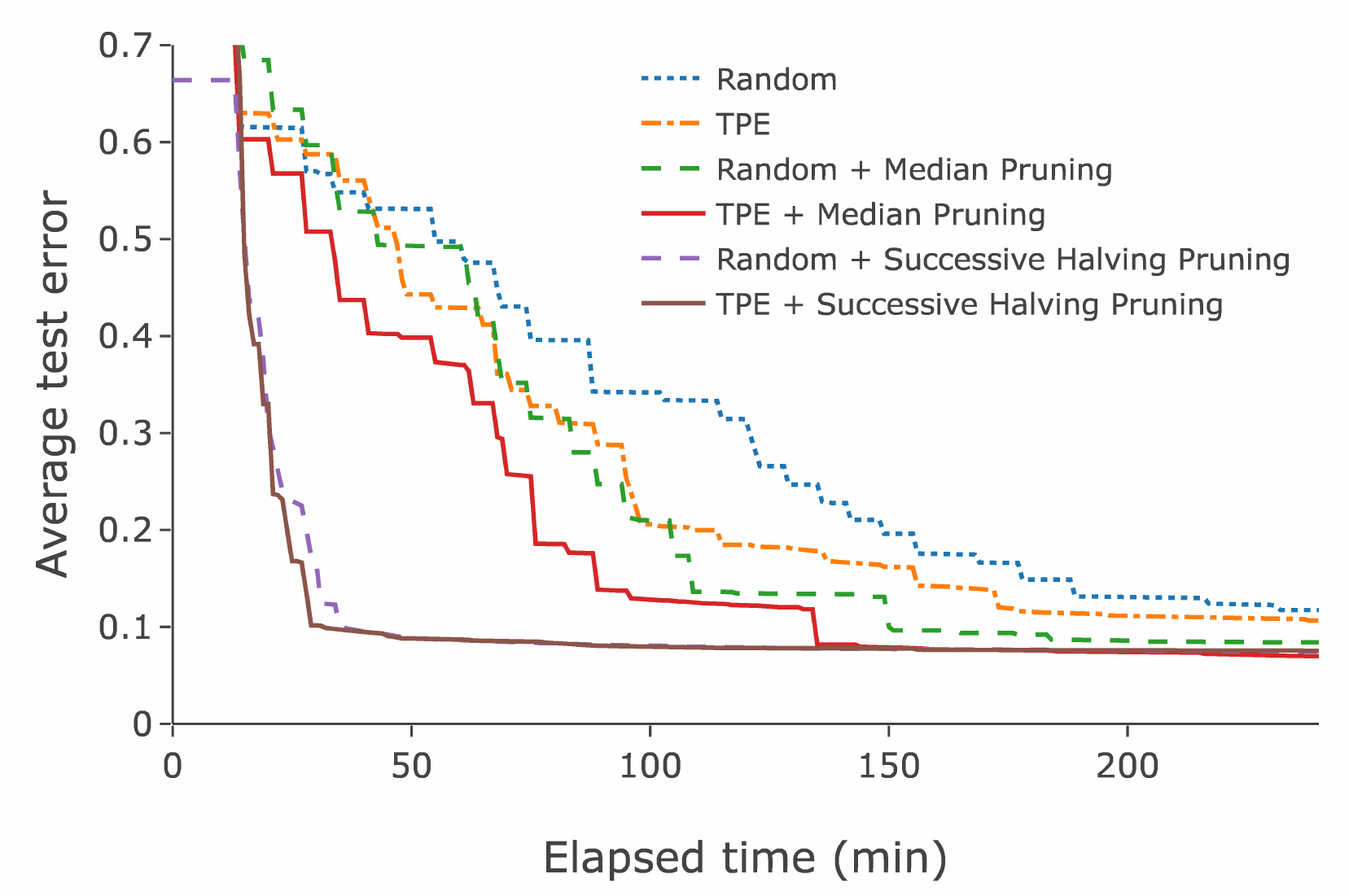}
\label{fig:alexnet_svhn}
\hspace{0.7em}
}
\subfloat[][]{
\includegraphics[width=6.0cm]{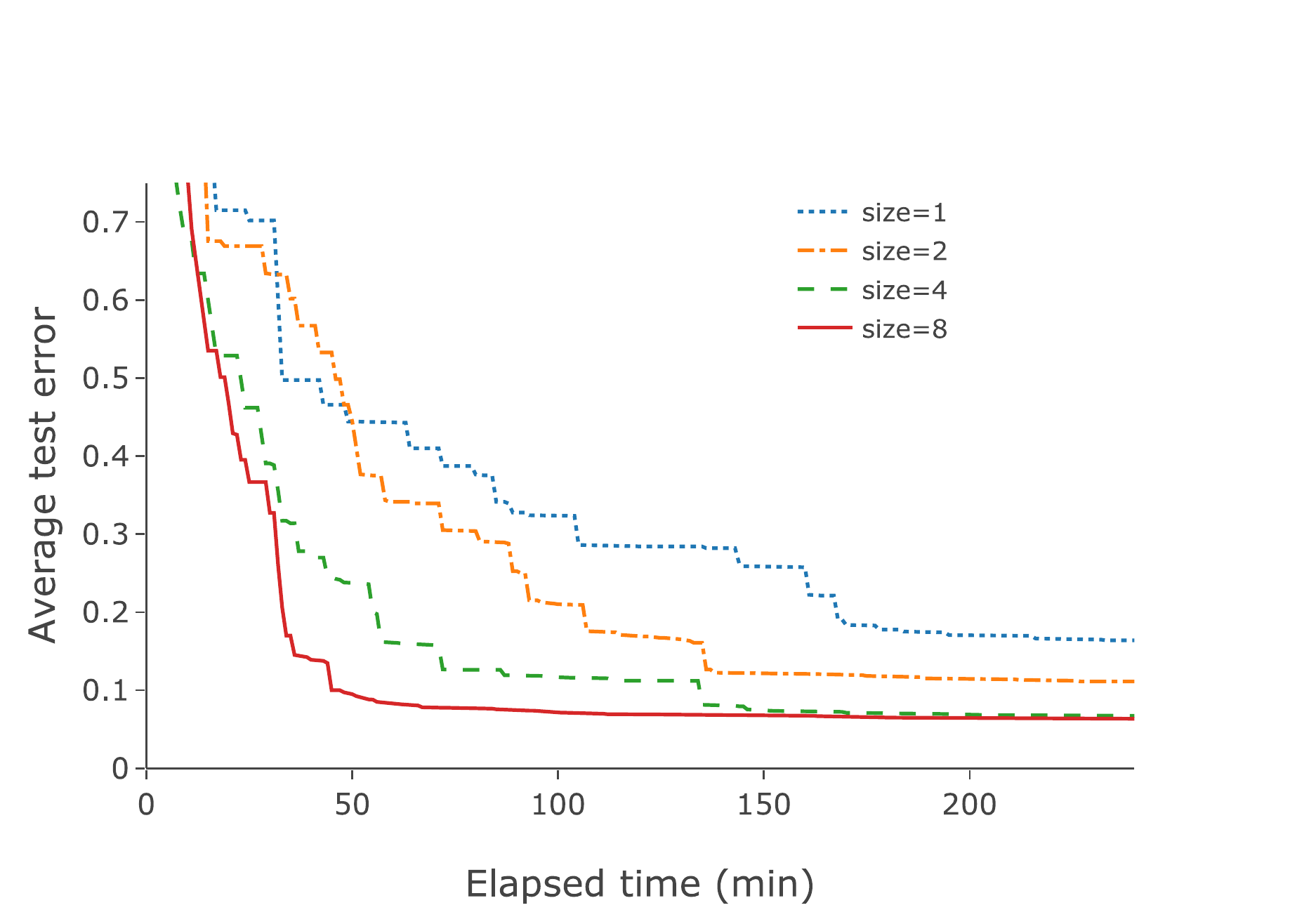}
\label{fig:distributed_alexnet_svhn_time}
\hspace{-0.7em}
}
\subfloat[][]{
\includegraphics[width=6.0cm]{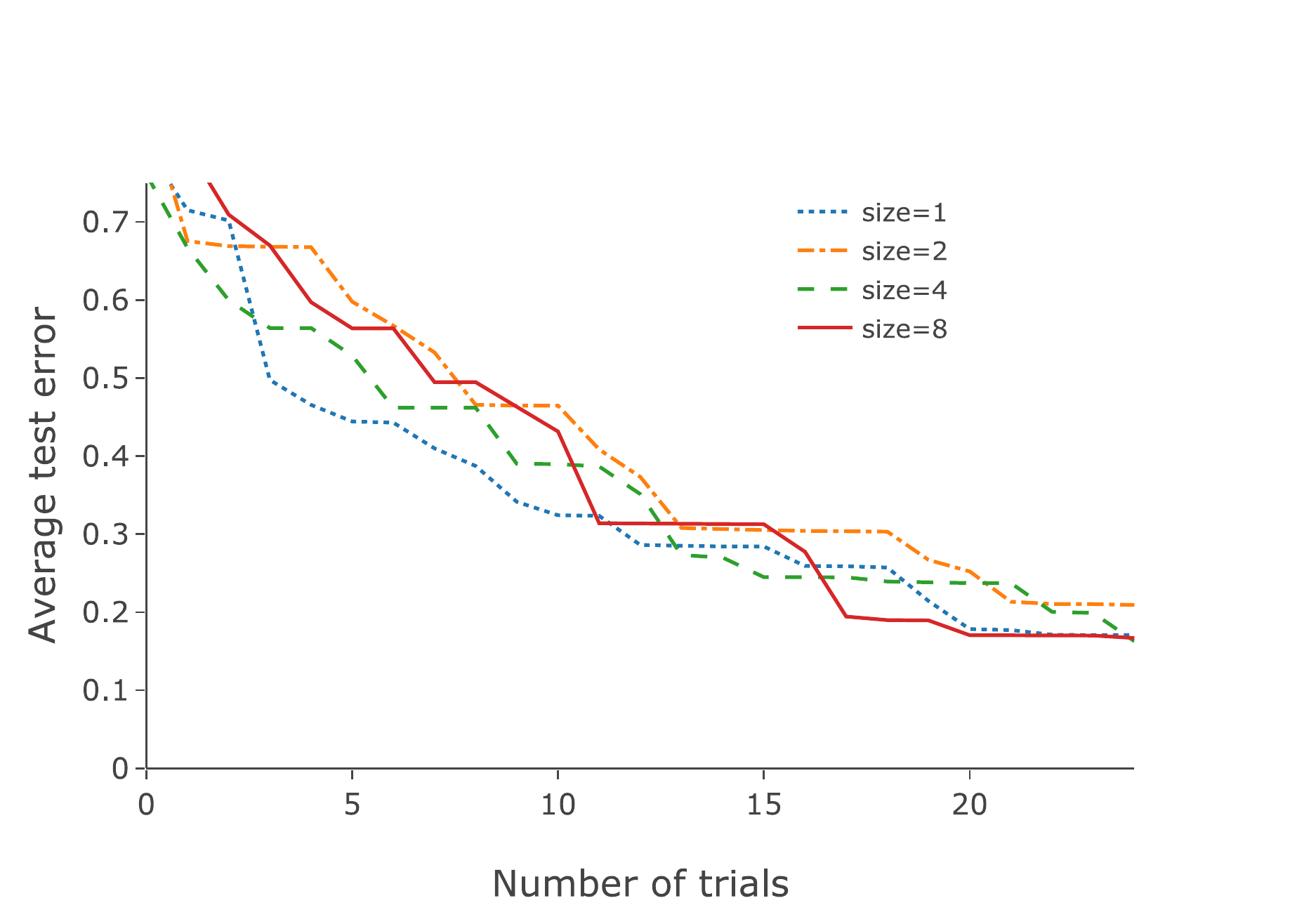}
\label{fig:distributed_alexnet_svhn_trials}
}
\caption{The transition of average test errors of simplified AlexNet for SVHN dataset. Figure (a) illustrates the effect of pruning mechanisms on TPE and random search. Figure (b) illustrates the effect of the number of workers on the performance. Figure (c) plots the test errors against the number of \emph{trials} for different number of workers.
Note that the number of workers has no effect on the relation between the number of executed \emph{trials} and the test error.
The result also shows the superiority of ASHA pruning over median pruning.}
\end{figure*}

We also evaluated the scalability of \emph{Optuna}'s distributed optimization.
Based on the same experimental setup used in Section~\ref{Performance Evaluation of Pruning}, we recorded the transition of the best scores obtained by TPE with 1, 2, 4, and 8 workers in a distributed environment.
Figure~\ref{fig:distributed_alexnet_svhn_time} shows the relationship between optimization score and execution time.
We can see that the convergence speed increases with the number of workers.

\begin{figure}
	  \begin{center}
			\includegraphics[width=6.5cm]{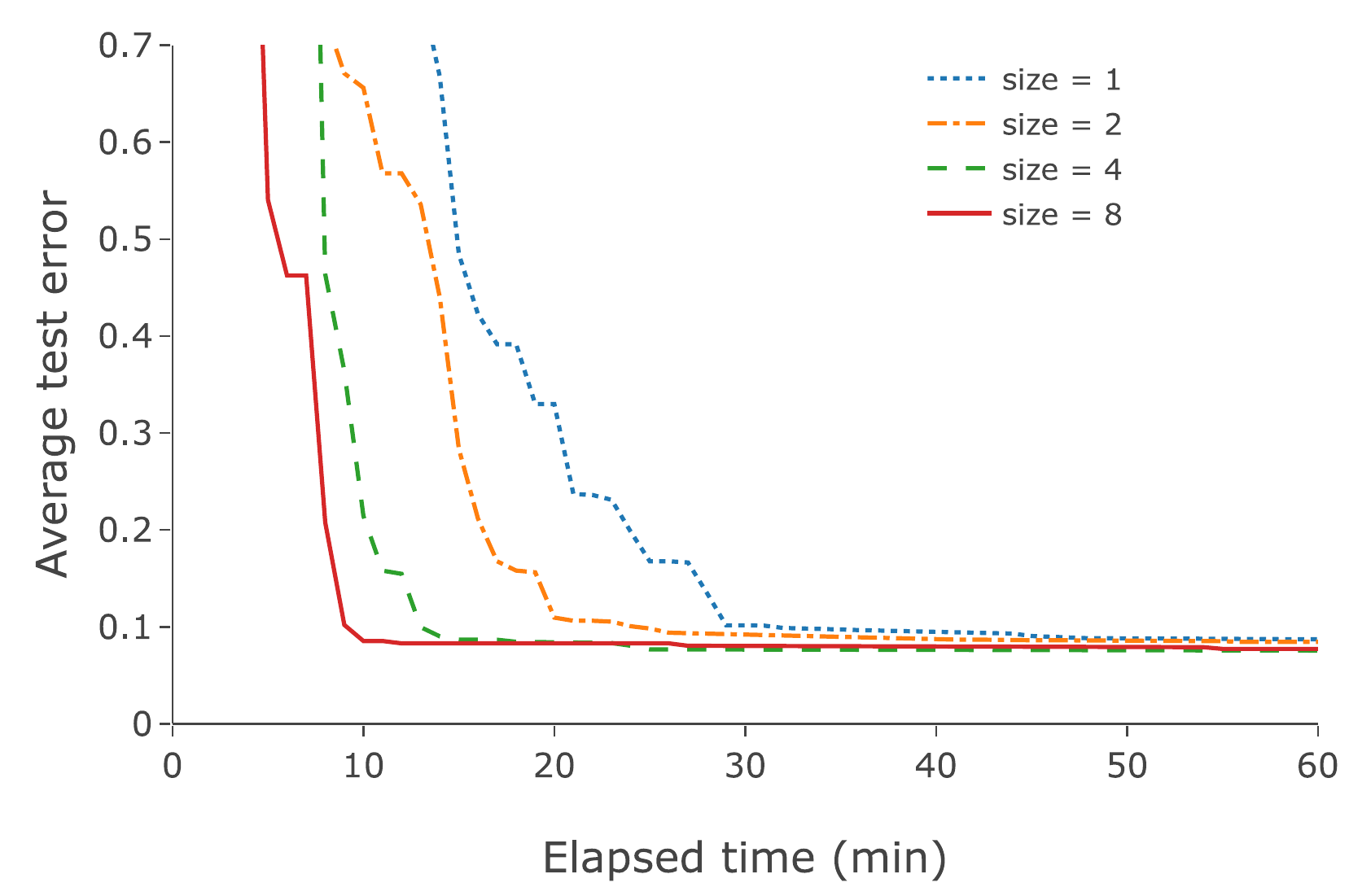}
			\caption{Distributed hyperparameter optimization process for the minimization of average test errors of simplified AlexNet for SVHN dataset.
			The optimization was done with ASHA pruning.}
	        \label{fig:pruning_distributed_alexnet_svhn_trials}
	  \end{center}
\end{figure}

In the interpretation of this experimental results, however,  we have to give a consideration to the fact that the relationship between the number of workers and the efficiency of optimization is not as intuitive as the relationship between the number of workers and the number of \emph{trials}.
This is especially the case for a SMBO~\cite{Hutter:2011:SMO:2177360.2177404} such as TPE, where the algorithm is designed to sequentially evaluate each \emph{trial}.
The result illustrated in Figure~\ref{fig:distributed_alexnet_svhn_trials} resolves this concern.
Note that the optimization scores per the number of \emph{trials} (i.e., parallelization efficiency) barely changes with the number of workers.
This shows that the performance is linearly scaling with the number of \emph{trials}, and hence with the number of workers.
Figure \ref{fig:pruning_distributed_alexnet_svhn_trials} illustrates the result of optimization that uses both parallel computation and pruning.
The result suggests that our optimization scales linearly with the number of workers even when implemented with a pruning algorithm.

%% file: sections/09-case_study.tex
\section{Real World Applications}
\emph{Optuna} is already in production use, and it has been successfully applied to a number of real world applications.
\emph{Optuna} is also being actively used by third parties for various purposes, including projects based on \emph{TensorFlow} and \emph{PyTorch}.
Some projects use \emph{Optuna} as a part of pipeline for machine-learning framework (e.g., \emph{redshells}\footnote{\url{https://github.com/m3dev/redshells}}, \emph{pyannote-pipeline}\footnote{\url{https://github.com/pyannote/pyannote-pipeline}}).
In this section, we present the examples of \emph{Optuna}'s applications in the projects at Preferred Networks.

\myparagraph{Open Images Object Detection Track 2018}
\emph{Optuna} was a key player in the development of Preferred Networks' Faster-RCNN models for Google AI Open Images Object Detection Track 2018 on Kaggle~\footnote{\url{https://www.kaggle.com/c/google-ai-open-images-object-detection-track}}, whose dataset is at present the largest in the field of object detection~\cite{openimages}.
Our final model, PFDet~\cite{Akiba2018PFDet2P}, won the 2nd place in the competition.

\vspace{0.5em}

As a versatile next generation optimization software, \emph{Optuna} can be used in applications outside the field of machine learning as well.
Followings are applications of \emph{Optuna} for non-machine learning tasks.

\myparagraph{High Performance Linpack for TOP500}
The \emph{Linpack} benchmark is a task whose purpose is to measure the floating point computation power of a system in which the system is asked to solve a dense matrix LU factorization.
The performance on this task is used as a measure of sheer computing power of a system and is used to rank the supercomputers in the TOP500 list\footnote{\url{https://www.top500.org/}}.
\emph{High Performance Linpack} (HPL) is one of the implementations for Linpack.
HPL involves many hyperparameters, and the performance result of any system heavily relies on them.
We used \emph{Optuna} to optimize these hyperparameters in the evaluation of the maximum performance of \emph{MN-1b}, an in-house supercomputer owned by Preferred Networks.

\begin{sloppypar}
\myparagraph{\emph{RocksDB}}
\emph{RocksDB}~\cite{dong2017optimizing} is a persistent key-value store for fast storage that has over hundred user-customizable parameters.
As described by the developers in the official website, "configuring \emph{RocksDB} optimally is not trivial", and even the "\emph{RocksDB} developers don't fully understand the effect of each configuration change"\footnote{\url{https://github.com/facebook/rocksdb/wiki/RocksDB-Tuning-Guide\#final-thoughts}}.
For this experiment, we prepared a set of 500,000 files of size 10KB each, and used \emph{Optuna} to look for parameter-set that minimizes the computation time required for applying a certain set of operations(store, search, delete) to this file set.
Out of over hundred customizable parameters, we used \emph{Optuna} to explore the space of 34 parameters.
With the default parameter setting, \emph{RocksDB} takes 372seconds on HDD to apply the set of operation to the file set.
With pruning, \emph{Optuna} was able to find a parameter-set that reduces the computation time to 30 seconds.
Within the same 4 hours, the algorithm with pruning explores 937 sets of parameters  while the algorithm without pruning only explores 39.
When we disable the time-out option for the  evaluation process,
the algorithm without pruning explores only 2 \emph{trials}.
This experiment again verifies the crucial role of pruning.
\end{sloppypar}

\myparagraph{Encoder Parameters for \emph{FFmpeg}}
\emph{FFmpeg}\footnote{\url{https://www.ffmpeg.org/}} is a multimedia framework that is widely used in the world for decoding, encoding and streaming of movies and audio dataset.
\emph{FFmpeg} has numerous customizable parameters for encoding.
However, finding of good encoding parameter-set for \emph{FFmpeg} is a nontrivial task, as it requires expert knowledge of codec.
We used \emph{Optuna} to seek the encoding parameter-set that minimizes the reconstruction error for the Blender Open Movie Project's "Big Buck Bunny"\footnote{Blender Foundation | www.blender.org}.
\emph{Optuna} was able to find a parameter-set whose performance is on par with the second best parameter-set among the presets provided by the developers.

%% file: sections/10-conclusions.tex
\section{Conclusions}
\begin{sloppypar}
The efficacy of \emph{Optuna} strongly supports our claim that our new design criteria for next generation optimization frameworks are worth adopting in the development of future frameworks.
The \emph{define-by-run} principle enables the user to dynamically construct the search space in the way that has never been possible with previous hyperparameter tuning frameworks.
Combination of efficient searching and pruning algorithm greatly improves the cost effectiveness of optimization.
Finally, scalable and versatile design allows users of various types to deploy the frameworks for a wide variety of purposes.
As an open source software, \emph{Optuna} itself can also evolve even further as a next generation software by interacting with open source community.
It is our strong hope that the set of design techniques we developed for \emph{Optuna} will serve as a basis of other next generation optimization frameworks to be developed in the future.
\end{sloppypar}

{
\small
\paragraph{Acknowledgement.}
The authors thank
R.~Calland, S.~Tokui, H.~Maruyama,
K.~Fukuda, K.~Nakago, M.~Yoshikawa, M.~Abe,
H.~Imamura, and Y.~Kitamura
for valuable feedback and suggestion.
}